\documentclass{article}

\usepackage[preprint]{xatl_pre/neurips_2023}

\usepackage[utf8]{inputenc} 
\usepackage[T1]{fontenc}    
\usepackage{hyperref}       
\usepackage{url}            
\usepackage{booktabs}       
\usepackage{amsfonts}       
\usepackage{nicefrac}       
\usepackage{microtype}      
\usepackage{xcolor}         

\usepackage{array}
\usepackage{multirow}
\usepackage{amsmath}
\usepackage{amssymb}
\usepackage{graphicx}
\usepackage{enumerate}
\usepackage{diagbox}
\usepackage{float}
\usepackage{xspace}
\usepackage{makecell}
\usepackage{cleveref}
\usepackage{bbold}
\crefformat{section}{\S#2#1#3}
\crefformat{subsection}{\S#2#1#3}
\crefformat{subsubsection}{\S#2#1#3}

\title{Cross-Architecture Transfer Learning for Linear-Cost Inference Transformers}

\newcommand{\methodfull}{Cross-Architecture Transfer Learning\xspace}
\newcommand{\methodabbr}{XATL\xspace}

\author{Sehyun Choi \\
Department of CSE, HKUST\\
Hong Kong S.A.R, China \\
\texttt{schoiaj@cse.ust.hk} \\
}

\begin{document}

\maketitle

\begin{abstract}
    Recently, multiple architectures has been proposed to improve the efficiency of the Transformer Language Models through changing the design of the self-attention block to have a linear-cost inference (LCI). A notable approach in this realm is the State-Space Machines (SSMs) architecture, which showed on-par performance on language modeling tasks with the self-attention transformers. However, such an architectural change requires a full pretraining of the weights from scratch, which incurs a huge cost to researchers and practitioners who want to use the new architectures. In the more traditional linear attention works, it has been proposed to approximate full attention with linear attention by swap-and-finetune framework~\citep{kasai-etal-2021-finetuning}. Motivated by this approach, we propose \methodfull (\methodabbr), in which the weights of the shared components between LCI and self-attention-based transformers, such as layernorms, MLPs, input/output embeddings, are directly transferred to the new architecture from already pre-trained model parameters. We experimented the efficacy of the method on varying sizes and alternative attention architectures and show that \methodabbr significantly reduces the training time up to 2.5x times and converges to a better minimum with up to 2.6\% stronger model on the LM benchmarks within the same compute budget.\footnote{The code and instructions to download the weights are published at \url{https://github.com/syncdoth/lit_llm_train}.}
\end{abstract}

\section{Introduction}

The NLP community is being flooded with multiple LLMs released everyday, ranging from large, closed-source, API-based LMs such as OpenAI GPT4~\citep{openai2023gpt4} \& ChatGPT~\citep{chatgpt}, Google's PaLM~\citep{chowdhery2022palm}, Anthropic AI's Claude~\citep{claude}, to small, open-source LMs like GPT~\citep{radford2019language}, LLaMA 1\&2~\citep{touvron2023llama, touvron2023llama2}, Mistral-7B~\citep{jiang2023mistral}, or Flacon-7B~\citep{falcon40b}. All of these models are built on some variant of the decoder stacks from Transformer~\citep{vaswani2017attention}, a self-attention based architecture with causal masking. This model architecture was proposed as a successor of Recurrent Neural Networks (RNNs), as it is highly parallelizable during training to allow efficient training and scalable to larger sizes by stacking layers. Still, one advantage of RNN over Transformer is its inference efficiency; RNN stores single global cache of previous states and need not re-compute softmax over the entire sequence, leading to constant time memory and operations in each generation step. On the other hand, in Transformers, the self-attention needs to maintain a full history of past KV-caches to accurately compute the self-attention for the current step, leading to quadratic inference time.

Recently, there is a new wave of exciting researches that propose new architectures addressing the inefficiency of self-attention's quadratic time cost during inference. It includes works in the line of linear attention transformers, which proposes kernels to approximate full softmax attention in linear time~\citep{katharopoulos_et_al_2020, qin-etal-2022-devil,  kitaev2020reformer, wang2020linformer, zhai2021attention, zhang2024hedgehog, arora2024simple}. Other approaches, often grouped as State-Space Machines, proposes new time-mixing methods that can be computed efficiently during both training and test time. For instance,  RWKV~\citep{peng2023rwkv}, RetNet~\citep{sun2023retentive}, H3~\citep{fu2023hungry}, Hyena~\citep{poli2023hyena} proposed methods that has both global and recurrent view, enabling parallel mode training and recurrent inferences. On the other hand, Mamba~\citep{gu2023mamba} introduced efficient scan algorithms and kernels to make long sequence training feasible and retain a recurrent view for efficient inference. We categorize these approaches as Low-Cost Inference Transformers (LCI).


%
While these architectures show strong promises of fast inference, they often require pre-training the models from scratch, which imposes a significant amount of compute requirement to the practitioners. This may hinder the wider adoption of such models, as the cost saved from efficient inference will have to be invested to the pre-training cost. Inspired by the Transformer-to-RNN (T2R, \citep{kasai-etal-2021-finetuning}) framework, which was originally proposed for linear attention kernels, we propose to address the costly pre-training requirement with a transfer learning approach where the weights are directly transferred to the new architecutre. In this framework named \methodfull (\methodabbr), the transformer architecture and the associated weights are kept the same, while the attention layers are swapped with different efficient components and trained on the language pre-training tasks. Most LCI and Transformer variants share core components, such as layernorms, residual connections, and stack of interleaved time-mixing (Attention) + channel-mixing (FFN) blocks. As the LCIs are essentially proposing a different method for time-mixing component, \methodabbr can transfer majority of the weights and enjoy strong initialization and accelerated training. Our experiments on popular LCI architectures and model sizes show that \methodabbr can achieve the same level of performance with 2.5x less compute and reach up to 2.6\% higher performance on language modeling and commonsense benchmarks with the same compute budget.

\section{Related Work}

\subsection{Linear Attention}

The works under the linear attention category proposes methods to approximate the softmax based attention with various kernels~\citep{katharopoulos_et_al_2020, qin-etal-2022-devil,  kitaev2020reformer, wang2020linformer, zhai2021attention}. The research focus is to design a kernel or method that can approximate the full expressivity of attention~\citep{zhang2024hedgehog, kasai-etal-2021-finetuning}, and propose to swap-then-finetune the linear attention only (T2R), or set the attention matrix as direct learning target and distill the attention matrix first. These methods therefore did not require pretraining from scratch and could substitute attention directly. However, they fail to retain local granularity when it comes to associative memory recall tasks~\citep{arora2023zoology}, and may approximate the attention too smoothly when the attention matrix can actually be characterized as ``spiky''~\citep{zhang2024hedgehog}.

\subsection{State Space Machines (SSMs)}

Recently, a new suite of models under the term Space Space Machines (SSMs) were introduced. Popular models include RWKV~\citep{peng2023rwkv}, RetNet~\citep{sun2023retentive}, H3~\citep{fu2023hungry}, Hyena~\citep{poli2023hyena}, and Mamba~\citep{gu2023mamba}. These blocks typically allow both parallel and recurrent view of the time mixing algorithm, allowing for fast parallel training and fast recurrent inference. In a practical standpoint, these models show great promise as their language model benchmark performances is on par with the State-of-the-Art Transformer models while being much efficient at inference time. However, they have new components (SSM Blocks) in its architecture, they must be pre-trained on a large-scale dataset first to be accessible by the general public. This work tries to address the expensive requirement of such pre-training by reusing the shared components' weights from a well-established pretrained transformer parameters.

\subsection{Weight Reusing}

We were also inspired by the Weight Reusing work that proposes to clone the weights of another model, such as in DistillBert~\citep{sanh2020distilbert}, and Weight Subcloning~\citep{samragh2023weight}. In DistillBert, the student model's weights are initialized from the teacher (Bert-base) model's weights, and Weight Subcloning proposes to clone the important weights from the larger model to a smaller model, which accelerates the training significantly. We also propose to initialize the weights of the new model from a pretrained-model, but not from a larger model but a same sized model with different attention architecture.

\section{\methodfull (\methodabbr)}

Inspecting various open-source LMs available, we noticed that many of the models share the same hidden dimension size as the other same-sized LMs. Moreover, Linear-Cost Inference (LCI) architectures borrow most of the components from the transformer architecture and work as a plug-and-play replacement to the self-attention block. Inspired by this, we propose a paradigm for transfer learning, called \methodfull (\methodabbr), which transfers weight matrices from other Pre-Trained Language Mdoels (PTLMs) to other language models with different architecture, especially the LCI architectures.

We first formalize the Transformer architecture in the equations below:

\begin{align}
    H^{(0)} &= E_i(X), \\
    H^{(l)} &= Block_l(H^{(l-1)}), \quad (\forall l \in \{1..L\}) \\
    Y &= E_o(LN(H^{(L)})),
\end{align}

where $X$ is input tokens, $E_i, E_o$ are input and output token embeddings, $Block_l$ is the $l$'th layer transformer block, $LN$ is layernorm, and $Y$ is the output logits. Each block is described as:

\begin{align}
    A &= Att(LN_1(H^{(l-1)})), \\
    O &= W_O(A) + H^{(l-1)}, \\
    H^{(l)} &= FFN(LN_2(O)) + O.
\end{align}

Notice that above formalization uses Pre-Normalization, which became de-facto in many SOTA LLMs.

We experimented with transferring the weights of token embedding ($E_i$) and LM head ($E_o$) layers, the FFN layers, and the attention output projection ($W_O$) weights. The layernorm weights are considered as the part of succeeding weights; for instance, $LN_2$ is copied along with FFN weights.

First, the token embeddings and the LM head layers are the very first and last layers and calibration of the weights will have direct impact on final loss computation. Also, they are the largest weight matrices, so initializing them to a well-learned weights may bring a significant benefit. Intuitively, copying the input token embedding can equip the model with a already good representation of the tokens to work with. Copying the LM head on the other hand provides the model a well-calibrated output matrix, and lets the model to learn a good representation that is compatible with that weight.

Next, the FFN layers has been interpreted to serve as the key-value memory storage for the models~\citep{geva-etal-2021-transformer}. Transferring these layers from a PTLM can be conceptually thought of as transferring the memory of the PTLMs to the new models. Moreover, they take up the most weights in the transformer architecture, and being able to initialize them to a good learned weight can potentially lead to accelerated training. The $W_O$ layer is the final projection layer of the self-attention block, and it can be considered as another smaller FFN applied after time-mixing.

\subsection{Freezing / Unfreezing of the Copied weights}
As the training progress, having a significant portion of weights frozen will be a disadvantage in terms of tunable parameters. To alleviate this problem, we propose a simple heuristic to schedule the unfreezing of the copied weights. We also experimented with unfreezing the weights from the beginning, which is similar to the T2R's swap-then-finetune method.

We monitored the improvement ratio of average loss between certain interval, and unfreeze the weights if the ratio is below some threshold. We also set patience parameter, to ensure that the threshold has been passed not due to randomness in the loss plane but because of saturation. In our experiments, the threshold was set to 1\% and the patience was set to 1. We refer to this method as \methodabbr-LIT, short for Loss Improvement Threshold.



\subsection{Hybrid Architecture}

Works like H3~\citep{fu2023hungry} and Hyena~\citep{poli2023hyena} suggests that Hybrid of attention and SSM blocks can boost the performance by a big margin. Moreover, in our case, the attention layers can also be copied over, allowing for more weights to be transferred. We followed the method used in the H3 paper and interleaved 2 attention layers with the SSM layers, at layer 2 and layer $N/2$ for a model with $N$ layers.

\section{Experiments}

In this section, we will describe the details of our experimental design and analyze the results. We will first introduce the models used (\cref{sec:models}), the datasets (\cref{sec:datasets}), the implementation detail (\cref{sec:implementation_detail}), and finally the analysis of the results (\cref{sec:results}).

\subsection{Models}
\label{sec:models}

For our main experiments, we compared the Multi-Head Attention (MHA) baseline with famous LCI architectures: RetNet~\citep{sun2023retentive}, and Mamba~\citep{gu2023mamba}. For the base Pre-Trained Language Model (PTLM), we used the Pythia-410m\footnote{Pretrained weights found at: \url{https://huggingface.co/EleutherAI/pythia-410m-deduped}}~\citep{biderman2023pythia} model's 100k checkpoint, which is trained on 200B tokens of the deduplicated Pile~\citep{gao2020pile} dataset. Note that this model is based on the GPTNeoX~\citep{black2022gptneox20b} architecture, and uses parallel residual connections. The LCI architectures may deviate from their original proposed architectures for FFN or residual connections to be compatible with this architecture. As the biggest difference comes from the SSM blocks or the linear attention blocks, we assumed these minor architectural differences to be less significant. Moreover, in the case of Mamba, we interleaved the Mamba blocks with FFN blocks, as opposed to homogeneous architecture of the original Mamba paper. This makes them compatible for weight transfer from Transformers. We name this model as StripedMamba.

\subsection{Datasets}
\label{sec:datasets}

Since we are transferring the weights from a PTLM, we trained the LCI models on the pretraining dataset used to train the PTLM that it was transferred from. As the main PTLM used was Pythia-410m model, we used the deduplicated Pile~\citep{gao2020pile} dataset\footnote{Dataset found at: \url{https://huggingface.co/datasets/EleutherAI/the_pile_deduplicated}}.

\subsection{Implementation Detail}
\label{sec:implementation_detail}

\begin{table}[ht]
    \small
    \centering
    \begin{tabular}{c|cccccc}
    \toprule
    Model & Attention & Hidden Dim. & Intermediate Dim. & Layers & Num. Heads \\
    \midrule
    Pythia-410m & MHA & 1024 & 4096 & 24 & 16 \\
    RetNet-430m & retention & 1024 & 4096 & 24 & 4 \\
    StripedMamba-430m & Mamba & 1024 & 4096 & 24 &  \\
    Pythia-1B & MHA & 2048 & 8192 & 16 & 8 \\
    RetNet-1B & retention & 2048 & 8192 & 16 & 8 \\
    \bottomrule
    \end{tabular}
    \caption{Model size configurations.}
    \label{tab:config}
\end{table}

We focused on two popular architectures: RetNet and (Striped-)Mamba. The model configurations follow the original pythia-410m for the most part (Table~\ref{tab:config}), except for the detailed configuration of the each LCI block. For RetNet, the head size was set to 256, following the suggestion of the original paper~\citep{sun2023retentive}, and for StripedMamba, the state dim, convolutional dim, and expand factor follow the original configuration found in the paper~\citep{gu2023mamba}.

We trained the 410m-sized models with Distributed Data Parallel using the Pytorch-Lightning~\citep{Falcon_PyTorch_Lightning_2019} framework and litgpt~\citep{litgpt-2023} codebase. We used flash-attention v2~\citep{dao2023flashattention2} for MHA and Hybrid models. For 1B models, they were trained with Fully Sharded Data Parallel strategy.

We used AdamW optimizer with betas 0.9, 0.95. The learning rate was set to 3e-4, with cosine annealing scheduler with minimum learning rate at 3e-5 and warmup steps of 2000. The Weight decay was set to 0.1. We used a batch containing 1024 examples, each spanning 2048 tokens. The models were trained with the same schedule as the pythia paper but stopped early at 150 billion training tokens, which translates to 75,000 steps. The training took 3 days on 32 NVIDIA A100 80GB PCie GPUs. For 1B, in the interest of time, they were stopped at 100B tokens.

\subsection{Results}
\label{sec:results}

In this section, we will analyze the empirical performance of models trained with the proposed \methodabbr training on various LM benchmarks. The benchmarks include language modeling benchmarks such as Lambada~\citep{paperno2016lambada} and commonsense reasoning benchmarks, namely HellaSwag~\citep{zellers2019hellaswag}, PIQA~\citep{bisk2019piqa}, ARC Challenge~\citep{yadav-etal-2019-quick}, and WinoGrande~\citep{sakaguchi2019winogrande}. We follow the standard procedures and report the accuracy of the models on the corresponding benchmarks (for LAMBADA, we also report the perplexity in Table~\ref{tab:benchmark}).

\subsection{Effect of Weight Transfer}
\begin{table}
    \fontsize{6}{7}\selectfont
    \renewcommand\arraystretch{1.2}
    \centering
    \begin{tabular}{c|cc|ccccccc}
    \toprule
        Size & Transferred Parameters & Model & LAMBADA & HellaSwag & PIQA & Arc-E & Arc-C & WinoGrande & Average \\
        &&& acc $\uparrow$ & acc $\uparrow$ & acc $\uparrow$  & acc $\uparrow$ & acc $\uparrow$ & acc $\uparrow$ & acc $\uparrow$ \\
    \midrule
    \multirow{12}{*}{400M} & \multirow{4}{*}{Scratch} & MHA & 47.43 & 38.23 & 65.89 & 49.62 & 23.72 & 51.07 & 45.99 \\
         && Mamba & 43.49 & 39.01 & 67.57 & 51.73 & 24.83 & 53.28 & 46.65 \\
         && RetNet & 44.63 & 36.67 & 65.29 & 50.13 & 24.4 & 50.36 & 45.25  \\
         && Hybrid RetNet & 46.21 & 37.83 & 66.32 & 50.29 & 25.51 & 53.43 & 46.6 \\
    \cline{2-10}
    &\multirow{2}{*}{$E_{i,o}$} & MHA & 47.23 & 38.27 & 66.27 & 49.79 & 23.29 & 51.85 & 46.12 \\
         && RetNet & 40.05 & 36.78 & 64.91 & 49.49 & 25.0 & 53.43 & 44.94 \\
    \cline{2-10}
         &\multirow{4}{*}{$+ FFN$} & MHA & 50.38 & 40.88 & 67.36 & 51.26 & 25.43 & 52.57 & 47.98 \\
         && Mamba & 49.72 & 41.66 & 67.19 & 52.1 & 24.32 & 53.51 & 48.08 \\
         && RetNet & 48.81 & 39.92 & 66.38 & 50.55 & 24.74 & 52.8 & 47.2 \\
         && Hybrid RetNet & 49.29 & 40.96 & 67.36 & 51.68 & 25.43 & 52.72 & 47.91 \\
    \cline{2-10}
         &\multirow{2}{*}{$ + W_O$} & MHA & 51.33 & 41.11 & 66.38 & 51.73 & 26.19 & 53.04 & 48.3 \\
         && RetNet & 47.8 & 39.72 & 66.7 & 50.51 & 25.51 & 51.54 & 46.96 \\
    \midrule
    \midrule
    \multirow{3}{*}{1B} & Scratch & RetNet & 49.3 & 41.46 & 66.86 & 52.98 & 25.63 & 52.82 & 48.18 \\
    \cline{2-10}
    & $E_{i,o} + FFN$ & RetNet & 51.47 & 44.97 & 69.26 & 54.5 & 25.51 & 51.14 & 49.48 \\
    & $E_{i,o} + FFN + ATTN$ & Hybrid RetNet & 53.66 & 45.29 & 69.7 & 54.92 & 27.56 & 53.43 & 50.76 \\
    \bottomrule
    \end{tabular}
    \caption{Ablation study on which parameter to transfer over multiple attention architectures. The pretrained weights originate from Pythia-410m model. 400M models were trained for 150B tokens (75k steps), while 1B models were trained for 100B tokens (50k steps).}
    \label{tab:transfer}
\end{table}

We first observe the effect of weight transfer on two model sizes, 400M and 1B, which are shown in Table~\ref{tab:transfer}. We can observe that for Multi-Head Attention (MHA), which serves as a baseline, shows consistent increase in performance as more weights are transferred. However, for RetNet, we could observe that transferring $W_O$ negatively affects performance. This might be because the retention outputs show strong differences from the trained self-attention outputs before normalization and therefore it was incompatible with $W_O$ weights. Also, copying the embeddings only did not prove to be beneficial, as the average performance did not surpass the RetNet model trained from scratch. Especially, the performance on the LAMBADA benchmark was 4.5 points below, while other tasks showed marginal improvements. For RetNet, the most beneficial \methodabbr configuration was to transfer the input/output embeddings and the FFN layers, and we used this configuration for other architectures.

For StripedMamba models, the first thing that can be noticed that it shows stronger performance than RetNet and even attention models, reaffirming the findings of the mamba paper. We could also observe a strong performance boost at the same number of tokens when the embeddings and FFN layers were transferred. Finally, for the Hybrid models, we could also observe a performance boost in all dimensions, except for Arc-Challenging subset and WinoGrande benchmark.

We also conducted experiments with 1B RetNet models, which were trained up to 100B tokens of Pile-Deduped dataset. We could observe a similar trend where the models with transferred weights from Pythia-1B shows stronger performance at the same number of tokens, and Hybrid models consistently outperforming the non-Hybrid counterpart.

\begin{figure}
\centering
    \includegraphics[width=0.4\linewidth]{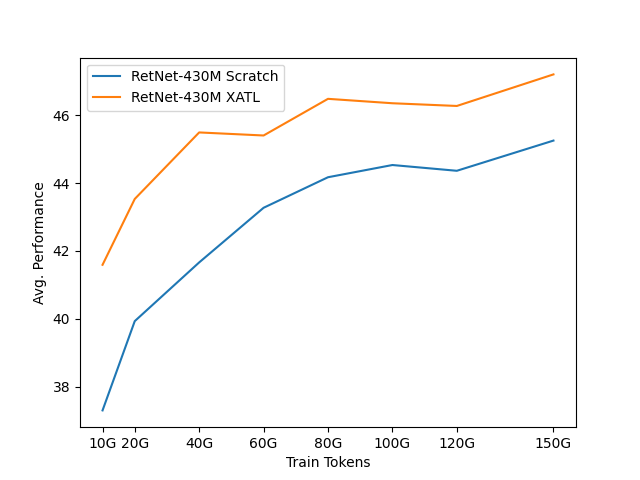}
    \includegraphics[width=0.4\linewidth]{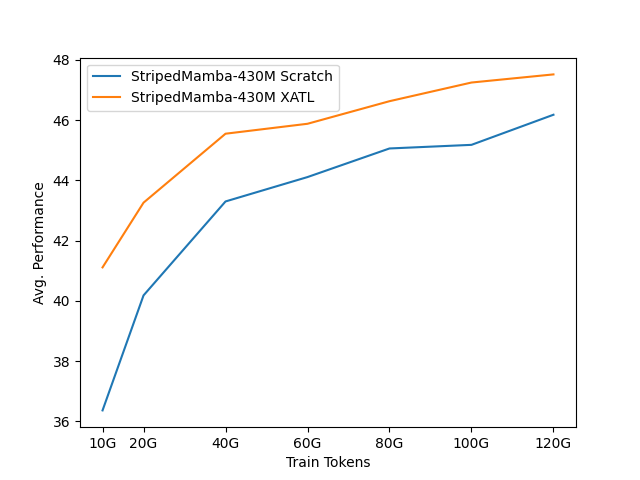}
    \caption{Comparison of average benchmark performance at different training checkpoints of RetNet-430M and StripedMamba-430M models.}
    \label{fig:early-training}
\end{figure}
Another advantage of \methodfull is that it boosts the initial training performance by a large margin, which can greatly help when the compute budget is limited.  In Figure~\ref{fig:early-training}, we show that \methodabbr training at 10B tokens is as performant as the 40B token checkpoint of the same model trained from scratch, and RetNet-430M-\methodabbr at 60B token already surpasses RetNet-430M Scratch model at 150B, suggesting a 2.5x reduced cost for same performance.

\subsection{Effect of Freezing}

\begin{table}
    \centering
    \fontsize{6}{7}\selectfont
    \renewcommand\arraystretch{1.2}
    \begin{tabular}{c|cc|ccccccc}
    \toprule
        Model & Transferred Parameters & Freeze & LAMBADA & HellaSwag & PIQA & Arc-E & Arc-C & WinoGrande & Average \\
        &&& acc $\uparrow$ & acc $\uparrow$ & acc $\uparrow$ & acc $\uparrow$ & acc $\uparrow$ & acc $\uparrow$ & acc $\uparrow$ \\
        \midrule
         \multirow{5}{*}{RetNet-430M} & \multirow{3}{*}{$E_{i,o} + FFN + W_o$} & Frozen & 33.55 & 35.52 & 64.31 & 48.23 & 24.15 & \textbf{52.96} & 43.12 \\
         & & Unfrozen & 47.8 & 39.72 & 66.7 & 50.51 & 25.51 & 51.54 & 46.96 \\
         & & LIT & 45.27 & 39.35 & \textbf{67.9} & \textbf{51.47} & \textbf{25.94} & 52.17 & 47.02 \\
         \cline{2-10}
         & \multirow{3}{*}{$E_{i,o} + FFN$} & Frozen & 39.78 & 37.66 & 65.67 & 49.62 & 23.98 & 50.59 & 44.55 \\
         & & Unfrozen & \textbf{48.81} & \textbf{39.92} & 66.38 & 50.55 & 24.74 & 52.8 & \textbf{47.2} \\
    \bottomrule
    \end{tabular}
    \caption{Freeze vs. Unfreeze vs. Scheduled Unfreeze strategies on RetNet models at both 400M. All weights are transfered from the corresponding pretrained weights. Trained for 150B tokens (75k steps).}
    \label{tab:freeze}
\end{table}

Next, we experimented with how the freezing of the transferred weight affect the training performance, and the results are shown in Table~\ref{tab:freeze}. LIT stands for Loss Improvement Treshold, where the transferred weights are frozen initially and unfrozen when the loss improvement compared to the previous checkpoint falls below a threshold. We found that in both configuration of weight transfer, freezing the weights affected negatively as the number of trainable weights are reduced towards the end, and LIT training to be marginally stronger than unfreezing the weights from the beginning. Hence, we simply chose to unfreeze the transferred weights, as they did not show much difference in performance empirically, and the implementation and training procedure is simpler.

\subsection{Comparison Against Open-Source Models}

\begin{table}
    \fontsize{6}{7}\selectfont
    \centering
    \renewcommand\arraystretch{1.2}
    \begin{tabular}{c|ccccccccc}
    \toprule
    \multirow{2}{*}{Model} & Tokens & LAMBADA & LAMBADA & HellaSwag & PIQA & Arc-E & Arc-C & WinoGrande & Average \\
    && ppl $\downarrow$ & acc $\uparrow$ & acc $\uparrow$ & acc $\uparrow$  & acc $\uparrow$  & acc $\uparrow$  & acc $\uparrow$ & acc $\uparrow$ \\
    \midrule
    Hybrid H3-360M & 300B & 89.48 & 25.77 & 31.7 & 64.2 & 44.4 & 24.2 & 50.6 & 40.1\\
    Pyhia-410M & 300B & 10.84 & 51.4 & 40.6 & 66.9 & 52.1 & 24.6 & 53.8 & 48.2 \\
    Mamba-370M & 300B & 8.14 & 55.6 & 46.5 & 69.5 & 55.1 & 28.0 & 55.3 & 50.0 \\
    \midrule
    Pyhia-410M & 150B & 11.38 & \textbf{49.99} & 40.18 & 66.87 & 50.55 & 24.06 & 53.51 & 47.53 \\
    Pyhia-410M-repr & 150B & 13.95 & 47.43 & 38.23 & 65.89 & 49.62 & 23.72 & 51.07 & 45.99 \\
    \underline{RetNet-430M-\methodabbr} & 150B & 12.24 & 48.81 & 39.92 & 66.38 & 50.55 & 24.74 & 52.8 & 47.2 \\  
    \underline{Hybrid RetNet-430M-\methodabbr} & 150B & 11.72 & 49.29 & 40.96 & \textbf{67.36} & 51.68 & \textbf{25.43} & 52.72 & 47.91 \\  
    \underline{StripedMamba-430M-\methodabbr} & 150B & \textbf{11.31} & 49.72 & \textbf{41.66} & 67.19 & \textbf{52.1} & 24.32 & \textbf{53.51} & \textbf{48.08} \\  
    \midrule
    \midrule
    Pythia-1B & 300B  & 7.92 & 56.1 & 47.2 & 70.7 & 57.0 & 27.1 & 53.5 & 51.9 \\
    Mamba-790M & 300B  & 6.02 & 62.7 & 55.1 & 72.1 & 61.2 & 29.5 & 56.1 & 57.1 \\
    \midrule
    Pythia-1B & 100B  & 9.24 & \textbf{54.3} & 42.02 & 68.99 & 53.83 & 25.68 & 52.09 & 49.49 \\
    TinyLlama-1.1B & 100B & 15.33 & 44.25 & 43.5 & 68.12 & 50.8 & 24.32 & 53.28 & 47.38 \\
    \underline{RetNet-1B-\methodabbr} & 100B  & 9.73 & 51.47 & 44.97 & 69.26 & 54.5 & 25.51 & 51.14 & 49.48 \\ 
    \underline{Hybrid RetNet-1B-\methodabbr} & 100B  & \textbf{9.02} & 53.66 & \textbf{45.29} & \textbf{69.7} & \textbf{54.92} & \textbf{27.56} & \textbf{53.43} & \textbf{50.76} \\ 
    \bottomrule
    \end{tabular}
    \caption{Comparison against other models with best models we have. The numbers for other models are taken from the mamba paper. \underline{Underlined} models are trained by us.
    Pythia-410M-repr means that these are checkpoints trained by us to reproduce the results of Pythia models, trained from scratch. In general, they perform slightly worse then the public checkpoints.}
    \label{tab:benchmark}
\end{table}

We compared the performance of models trained by us with the open-source LMs of similar sizes. The results are shown in Table~\ref{tab:benchmark}. The table is adapted from the Mamba paper and includes the performance of models trained to 300B tokens of the Pile dataset. As we did not train models to 300B tokens for our experiments, we included evaluation results from earlier checkpoints of the models found publicly. Also, for Pythia-410M models, we have reproduced the training from scratch in the previous experiments, and also included their performance (Pythia-410m-repr in the Table~\ref{tab:benchmark}).

We could observe that in the 400M model range at 150B tokens, \methodabbr can train RetNet models that are on par or stronger than transformers, especially when used in Hybrid with attention. This shows the strength of directly transferring compatible weights, as the same sized RetNet trained from scratch does not surpass the performance of Pythia-410M.

We could also observe a similar trend in the 1B sized model realm, with the \methodabbr version of RetNet model staying on par with the Pythia-1B model, and the Hybrid version surpassing the performance of the Pythia-1B model, from which the weights are copied from.

Although the RetNet-\methodabbr models did not surpass the performance of the Transformer (Pythia) even at 1B size, this is somewhat expected, as the original paper of the RetNet also reported a performance lower Transformer at 1B size~\citep{sun2023retentive}, which was re-affirmed by our reproduction of training from scratch in Table~\ref{tab:transfer}. We stress that \methodabbr consistently showed improvement of performance at the same number of tokens, which was what allowed for the RetNet models to stay on par with the Pythia models of the similar size.

\section{Limitations and Future Works}

Due to computational cost, the models were not overtrained up to 300B tokens, which would allow for full comparison with the other open source models. To make a fair comparison, we performed evaluation of the open source models with their public intermediate checkpoints. Also, it has been shown in the RetNet paper that the RetNet models start to surpass the performance of Transformer at 3B size. Moreover, the Pythia scaling suite~\citep{biderman2023pythia} observed a phase change phenomenon starting from 3B size, which together suggests that 3B and above is a interesting model size to experiment the effect of \methodabbr in a longer training scenario. We plan to carry out this direction after this version.

As the focus of this work was to show that the LCI models need not be entirely trained from scratch and either reduce the training time or improve the performance, we did not dive deep into the specifics of the LCI specific evaluations, such as measuring associative recall~\citep{arora2023zoology} or inference cost analysis. As this is not the main contribution of this work, we refer the readers to the original papers for this.

\section{Conclusion}

We proposed a weight transfer learning paradigm called \methodfull for LCI models, alleviating the need to train new architectures from scratch by transferring compatible components' weights from the pre-trained models. We have conducted various experiments regarding which components are the most effective when transferred, whether to freeze the transferred weights or not, and how does the \methodabbr help attention-hybrid architectures. We also compared the models trained with the best configurations with the open source models of similar sizes, and trained models that stays on par or outperforms similar sized SOTA models on the same number of training tokens, which was not possible when trained from scratch. \methodabbr has shown its effectiveness by significantly improving the performance of the LCI models at the same compute budget and reducing the pre-training compute to achieve the same performance as training from scratch.


\begin{ack}

We thank Nucleus AI for providing the compute and constructive feedbacks for conducting experiments and performing analysis.

\end{ack}

\bibliography{xatl_preprint}
\bibliographystyle{xatl_pre/iclr2024}


\end{document}